\documentclass[10pt,twocolumn,letterpaper]{article}
\usepackage{iccv}
\usepackage{times}
\usepackage{epsfig}
\usepackage{graphicx}
\usepackage{amsmath}
\usepackage{amssymb}

\usepackage{multirow}
\usepackage{makecell}
\usepackage{multicol}
\usepackage{mathrsfs}
\usepackage[ruled]{algorithm2e}
\usepackage{cellspace}

\usepackage{makecell}
\usepackage{bm}
\usepackage{paralist}
\usepackage{booktabs}
\usepackage[table]{xcolor}
\usepackage{colortbl}
\usepackage{bbding}

\usepackage[breaklinks=true,bookmarks=false]{hyperref}

\newcommand{\vpara}[1]{\vspace{0.07in}\noindent\textbf{#1 }}
\definecolor{demphcolor}{RGB}{144,144,144}
\newcommand{\tabincell}[2]{\begin{tabular}{@{}#1@{}}#2\end{tabular}}
\newcommand{\demph}[1]{\textcolor{demphcolor}{#1}}
\def\*#1{\bm{#1}}

\def\!#1{\vec{\bm{#1}}}

\usepackage[breaklinks=true,bookmarks=false]{hyperref}

\iccvfinalcopy 

\def\iccvPaperID{12736} 

\ificcvfinal\fi

\begin{document}

\newcommand{\yz}[1]{\textbf{\color{red}[(YZ: #1)]}}

\title{ViLTA: Enhancing Vision-Language Pre-training through Textual Augmentation}


\author{
Weihan Wang$^*$\qquad\quad Zhen Yang\thanks{Equal contribution.}\qquad\quad Bin Xu\thanks{Corresponding author.}\qquad\quad Juanzi Li\qquad\quad Yankui Sun 
\\[2mm]
Tsinghua University \
\\[1mm]
{\tt\small \{wangwh21, yangz21\}@mails.tsinghua.edu.cn\qquad\quad \{xubin, lijuanzi, syk\}@tsinghua.edu.cn  }
}


\newcommand{\model}{ViLTA}
\newcommand{\smodel}{\model\space}

\maketitle
\ificcvfinal\thispagestyle{empty}\fi

\begin{abstract}
Vision-language pre-training (VLP) methods are blossoming recently, and its crucial goal is to jointly learn visual and textual features via a transformer-based architecture, demonstrating promising improvements on a variety of vision-language tasks. Prior arts usually focus on how to align visual and textual features, but  strategies for improving the robustness of model and speeding up model convergence are left insufficiently explored.

In this paper, we propose a novel method \model, comprising of two components to further facilitate the model to learn fine-grained representations among image-text pairs. For Masked Language Modeling (MLM), we propose a cross-distillation method to generate soft labels to enhance the robustness of model, which alleviates the problem of treating synonyms of masked words as negative samples in one-hot labels. For Image-Text Matching (ITM), we leverage the current language encoder to synthesize hard negatives based on the context of language input, encouraging the model to learn high-quality representations by increasing the difficulty of the ITM task. By leveraging the above techniques, our \smodel can achieve better performance on various vision-language tasks. Extensive experiments on benchmark datasets demonstrate that the effectiveness of \smodel and its promising potential for vision-language pre-training.

\end{abstract}

\section{Introduction}\label{sec:intro}

\begin{figure*}
    \centering 
    \setlength{\belowcaptionskip}{-5pt}  \includegraphics[width=\textwidth]{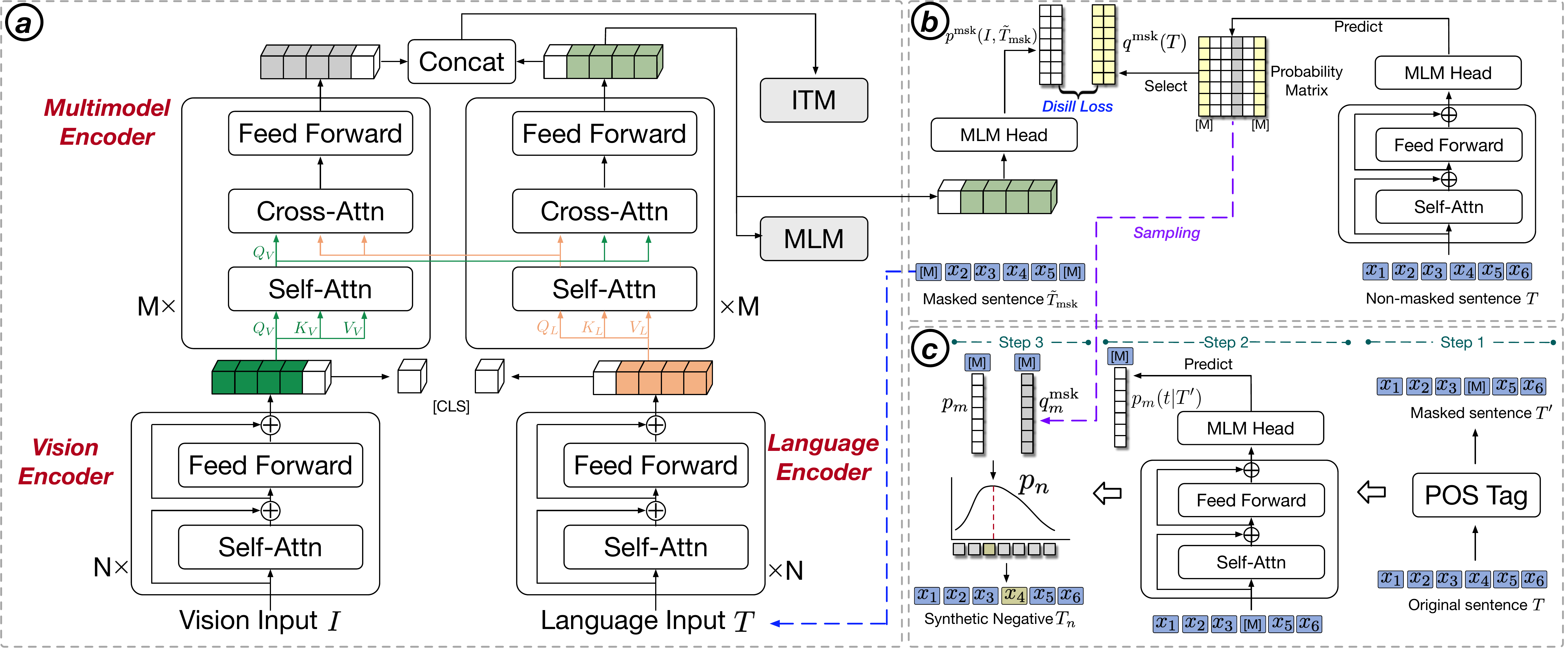}
    \caption{The overall architecture of \model. The framework of \smodel contains three components, including vision, language, and multimodal encoders (Cf. (a)); soft labels obtained by the froze language encoder to enhance the robustness of model with noisy data in MLM (Cf. (b)); synthetic hard negatives generated by the current language encoder for ITM (Cf. (c)).}
    \label{fig:model_architecture}
\end{figure*}

Recent advancements in Vision-Language Pre-training (VLP) have achieved significant improvements in a wide range of multimodal tasks, such as visual question answering (VQA)~\cite{antol2015vqa}, image captioning~\cite{anderson2018bottom}, and image-text retrieval~\cite{lin2014microsoft,ren2015faster}. The target of VLP generally starts with training a model on massive image-text pairs in a self-supervised way, which empowers a new paradigm for finetuning various downstream tasks. Most recent VLP models~\cite{radford2021learning,li2021align,bao2021vlmo,zhang2022glipv2,bao2021beit} usually utilize a transformer-based architecture~\cite{vaswani2017attention} with some specific training techniques (e.g. image-text contrastive learning (ITC), masked language modeling (MLM), and image-text matching (ITM)) to align visual and textual information. These models have achieved outstanding performance on a variety of multimodal benchmarks, which further advances the field of multimodal representation learning.   
However, the above VLP models also suffer from two vital limitations: (1) one-hot labels in MLM hinder the robustness of the model; (2) negative samples selection in ITM affects the model convergency and downstream performance.

In specific, the MLM task is designed to predict masked tokens in a given language input by utilizing both visual and textual features. However, compared to traditional MLM approaches in NLP~\cite{devlin2018bert}, the MLM task in vision-language pre-training presents a unique limitation. VLP models use a pre-trained language model for secondary pre-training on image-text pairs, which may result in a loss of knowledge initially acquired from NLP datasets. Empirical results from previous studies~\cite{dou2022empirical} indicate that pre-training on multimodal datasets may lead to degraded performance on unimodal language tasks. Moreover, the presence of multiple candidate words to fill a masked position in an image-text pair can hinder the training of MLM. For instance, in the sentence "Two giraffes pace around their habitat", substituting "pace" with "walk" does not alter the sentence's meaning. Consequently, treating these words as negative samples in one-hot labels can impede MLM training.

As a popular pre-training task in VLP, the goal of ITM task is to distinguish positive and negative image-text pairs based on the learned representations. The common and straightforward way for negative selection is to randomly sample negatives for any given image-text pair. However, such a simple method can not provide more contributions for model convergence and result in sub-optimal performance. As a result, the model can easily achieve high accuracy in the first training epoch, usually above 99\%. To further improve the model's ability to learn fine-grained representations, an effective method is to offer hard negative samples to make the pre-training task more challenging. These hard negative samples are similar to positives and it is difficult for the model to distinguish them from positives.  A growing body of research~\cite{shekhar2017foil,parcalabescu2021valse,hendricks2021probing} illustrates that mining hard negatives can drastically alter the performance of multimodal models among various tasks, highlighting the significance of hard negative samples for enhancing the representation capabilities of the model. However, the existing attempts in mining hard negatives for vision-language pre-training only focus on sampling negatives in the discrete data space, ignoring the relationship among image-text pairs and the context of language input. 

\vpara{Present Work.}
To address the abovementioned issues, we propose a novel vision-language pre-training model named \model, comprising two key components. For MLM, we propose a \textit{cross-distillation} method to generate soft labels for improving the learning efficiency and boosting the robustness of the model. In specific, such a distillation method leverages the frozen language encoder to generate soft labels, which can be integrated into the original MLM task for joint training. In ITM, we propose to synthesize hard negatives based on the current language model by leveraging the context of language input, which is significantly different from previous works that select hard negatives from the raw data~\cite{li2021align,bao2021vlmo}. By utilizing these two techniques, our proposed \smodel can achieve better performance on a variety of downstream tasks, including visual question answering, visual entailment, visual reasoning, image-text retrieval, and image captioning. Extensive experimental results demonstrate the effectiveness of \model.
We summarize the contributions of this work as follows: \\
\indent 1) The proposed knowledge distillation method \textit{cross-distillation} generates soft labels to allow the model to better capture representations among image-text pairs, enabling the learning of the model more smooth and robust. \\
\indent 2) As opposed to sampling hard negatives from the raw data, we propose a strategy to synthesize hard negative samples based on the current language model, boosting the representation ability of the model by enhancing the difficulty of the ITM task.  \\
\indent 3) By effectively integrating these two techniques, our \smodel brings about outstanding performance improvements on various downstream tasks, demonstrating the superiority of \model.

\section{Related Work}\label{sec:related}

\noindent \textbf{Vision-Language Representation Learning.}
In prior research, several techniques have been utilized to integrate visual and language features for multimodal learning. One approach is to use pre-trained object detectors as feature extractors. ViLBERT~\cite{lu2019vilbert} and LXMERT~\cite{tan2019lxmert} employed co-attention for modality fusion, where two independent transformer modules were used for visual and language features, respectively. 
Alternatively, VisualBERT~\cite{li2019visualbert}, VL-BERT~\cite{su2019vl}, UNITER~\cite{chen2020uniter}, OSCAR~\cite{li2020oscar}, VinVL~\cite{zhang2021vinvl}, and VL-T5~\cite{cho2021unifying} employed a merged-attention mechanism, where visual and language features were inputted directly into a single transformer module for information exchange.

However, due to the low computational efficiency of object detectors and the inability to update weights during multimodal pre-training, researchers have shifted towards end-to-end pre-trained models. CLIP-ViL~\cite{shen2021much} and Pixel-BERT~\cite{huang2020pixel} fed grid features extracted by CNNs and text features into a single transformer module. ViLT~\cite{kim2021vilt} concatenated image patch embeddings and text token embeddings for pre-training.
Recent works focus on unifying the structure of image and text encoders. For instance, ALBEF~\cite{li2021align}, METER~\cite{dou2022empirical}, Florence~\cite{yuan2021florence}, CoCa~\cite{yu2022coca}, Flamingo~\cite{alayrac2022flamingo}, and PaLI~\cite{chen2022pali} utilized ViT~\cite{dosovitskiy2020image} as an image encoder, thereby unifying the structure of image and text encoders to some extent. In contrast, VLMo~\cite{bao2021vlmo} and \textsc{BEiT}-3~\cite{wang2022image} adopted multi-way transformers to unify the modeling of text and images where text, images, and image-text pairs are fed into a single transformer module.

\noindent \textbf{Knowledge Distillation.}
Knowledge distillation~\cite{hinton2015distilling} is firstly proposed to transfer the knowledge learned by the teacher model to the student model. It has wide-ranging applications across various modalities~\cite{sanh2019distilbert,jiao2019tinybert,touvron2021training,liu2021kd,li2021align,wang2022cliptd,dai2022enabling} to reduce the number of parameters and improve the performance of the student model. Among these works, KD-VLP\cite{liu2021kd} proposed an object-aware end-to-end VLP framework with object knowledge distillation. CLIP-TD~\cite{wang2022cliptd} used CLIP-targeted distillation to distill knowledge from both CLIP’s vision and language branches
into the existing VL model. VLKD~\cite{dai2022enabling} aligned the CLIP text encoder and BART encoder to enable the capability for multimodal generation. The work is similar to our method in the multimodal field is ALBEF~\cite{li2021align}, which employed momentum distillation by using its own momentum model as the teacher model.
However, while these works have focused on knowledge transfer within the same modality, our proposed \textit{cross-distillation} method aims to transfer knowledge from a language model to a multimodal model.

\vpara{Negative Sampling.} 
Hard negative mining is a widely adopted technique to enhance model performance. Prior research~\cite{cai2020all} has demonstrated that hard negative samples have the most significant impact on the training process, rendering the easiest 95$\%$ of negatives redundant.~\cite{shekhar2017foil,parcalabescu2021valse} generated a substantial number of hard negative samples through adversarial word replacement, resulting in a significant decline in the performance of several multi-modal models. On the other hand, works such as~\cite{gupta2020contrastive,zhang2021understanding,xia2022progcl} employed hard negative mining to improve model performance in contrastive learning. For ITM tasks, recent works~\cite{li2021align,bao2021vlmo,dou2022coarse} employed hard negative mining by selecting negative samples with the highest cosine similarity to the positive samples in the same batch. Nevertheless, this method has two limitations: First, the selection of negative samples is significantly influenced by the learning of the ITC task and the batch size. Second, this method has a high probability of choosing false negative samples, thereby adversely affecting the model's learning. To the best of our knowledge, our proposed method is the first to employ self-generated synthetic hard negative samples to enhance performance in image-text matching tasks.

\section{Method}\label{sec:method}

To improve vision-language pre-training, we propose \smodel that comprises of two components: 1) Knowledge Distillation for MLM; 2) Synthetic Hard Negatives for ITM. Figure~\ref{fig:model_architecture} shows the overall model architecture of \model.

The goal of \smodel is to further improve downstream performance of vision-language models by leveraging textual augmentation. To achieve it, the first is to utilize the frozen language encoder to generate soft labels for MLM. The second is to provide hard negatives for ITM by synthesizing negatives based on the current language encoder, which is significantly different from previous negatives selection method~\cite{li2021align,bao2021vlmo}.

\subsection{Model Architecture}\label{subsec:architecture}
As shown in Figure~\ref{fig:model_architecture}, the overall model architecture comprises of three components, including vision encoder, language encoder, and multimodal encoder. Here, we introduce each component in detail. 

\vpara{Vision Encoder.}
We employ ViT~\cite{dosovitskiy2020image} as a vision encoder to model an input image, which directly feeds image patches segmented from a whole image input and encodes them as encodes them as a sequence of embeddings $\{v_{cls}, v_1, ..., v_N\}$ with a additional \texttt{[CLS]} token embedding. Following the success in previous works~\cite{shen2021much,dai2022enabling,dou2022empirical,li2022mplug}, we initialize the weights of the image encoder using a pre-trained CLIP-ViT-224/16 model \cite{radford2021learning}.

\vpara{Language Encoder.} We leverage RoBERTa~\cite{liu2019roberta} as a language encoder to model language inputs. It converts the input caption into a sequence vector $\{w_{cls}, w_1, ..., w_N\}$, in which the embedding of the \texttt{[CLS]} token summarizes the global text feature. This sequence vector is then fed into the subsequent multimodal encoder to explore the relationship between image and text pairs. 

\vpara{Multimodal Encoder.}
To further capture the relationship among image-text pairs, we adopt a multimodal encoder which employs two independent cross-attention transformer modules to deeply fuse image and text information. The cross-modality multi-head attention module uses the representations of one modality as the query and another modality's representations as the key and value, as shown in Figure~\ref{fig:model_architecture}. This deep fusion mechanism independently encodes image and text features and fuses cross-model interaction, leading to better performance improvements.

\subsection{Knowledge Distillation for MLM}

Masked Language Modeling (MLM) aims to predict masked words by leveraging the learned image and text features. In specific, for any certain image-text pair, we first randomly mask a portion of tokens in a sentence by substituting them with the special token \texttt{[MASK]}. Then, the original masked tokens can be predicted by the remaining text input $\tilde{T}_{\text{msk}}$ and its corresponding image input $I$. Thus, the MLM task can be formulated with a cross-entropy loss:
\begin{equation}
    \mathcal{L}_{\text{mlm}} = \mathbb{E}_{(I, \tilde{T}_{\text{msk}}) \sim D} H(y^{\text{msk}}, p^{\text{msk}}(I, \tilde{T}_{\text{msk}}))
    \label{eq:mlm_task}
\end{equation}
where $D$ represents the whole training image-text pairs, $p^{\text{msk}}(I, \tilde{T}_{\text{msk}})$ denotes the predicted probability of the masked token, and $y^{\text{msk}}$ is a one-hot representation of the randomly masked ground-truth token.

Different from the traditional single-modality text encoders~\cite{lan2019albert,liu2019roberta} and previous vision-language encoders~\cite{li2021align,li2020oscar} that randomly mask 15\% of the input tokens, we increase the mask ratio from 15\% to 50\% in order to encourage the model to reconstruct the masked token by leveraging on both the context of text and image features. Such a mask ratio in \textsc{BEiT}-3~\cite{wang2022image} also verifies that a higher mask ratio can urge the model to recover the masked token from the content of the image rather than depending only on the context of the text itself.

In the MLM task, if the model is trained to only learn one-hot labels and treat all other potential positive examples as negative examples, it could harm the model's ability to learn effectively. In this work, we propose a novel \textit{cross-distillation} method to generate soft labels that can replace one-hot labels. Specifically, we duplicate the language encoder as a teacher model and freeze its parameters. Next, we input the \textit{non-masked} language sequence into the frozen language encoder to obtain a predicted probability matrix of a sequence of tokens. This is achieved by adding a masked language modeling (MLM) head on top of the output embeddings. We denote the predicted probability matrix as $q(T) \in \mathbb{R}^{S \times V}$, where $S$ is the length of the text sequence and $V$ is the vocabulary size. Subsequently, we select the predicted probability vector $q^{\text{msk}}(T)$ of the original masked tokens from $q(T)$ and use KL-divergence to measure the difference between the prediction of the multimodal encoder $p^{\text{msk}}(I, \tilde{T}_{\text{msk}})$ and the soft labels $q^{\text{msk}}(T)$. The distillation loss for MLM is defined as follows:
\begin{equation}
    \mathcal{L}_{\text{dis}} = \mathbb{E}_{(I,\tilde{T}_{\text{msk}}) \sim D, T \sim D} KL(q^{\text{msk}}(T), p^{\text{msk}}(I, \tilde{T}_{\text{msk}})) 
    \label{eq:distill_loss}
\end{equation}

As a result, the final loss can be formulated by the combination of the original MLM loss and distill loss:  
\begin{equation}
\mathcal{L}_{\text{mlm}}^{\text{dis}} = \alpha \mathcal{L}_{\text{mlm}} + (1 - \alpha) \mathcal{L}_{\text{dis}}
\label{eq:all_mlm}
\end{equation}
where $\alpha$ is a hyperparameter that controls the distillation weight. In our experiments, we set $\alpha$ as 0.5 for simplicity.

\begin{figure}
    \centering 
    \setlength{\abovecaptionskip}{5pt}
    \setlength{\belowcaptionskip}{-5pt}  \includegraphics[width=0.48\textwidth]{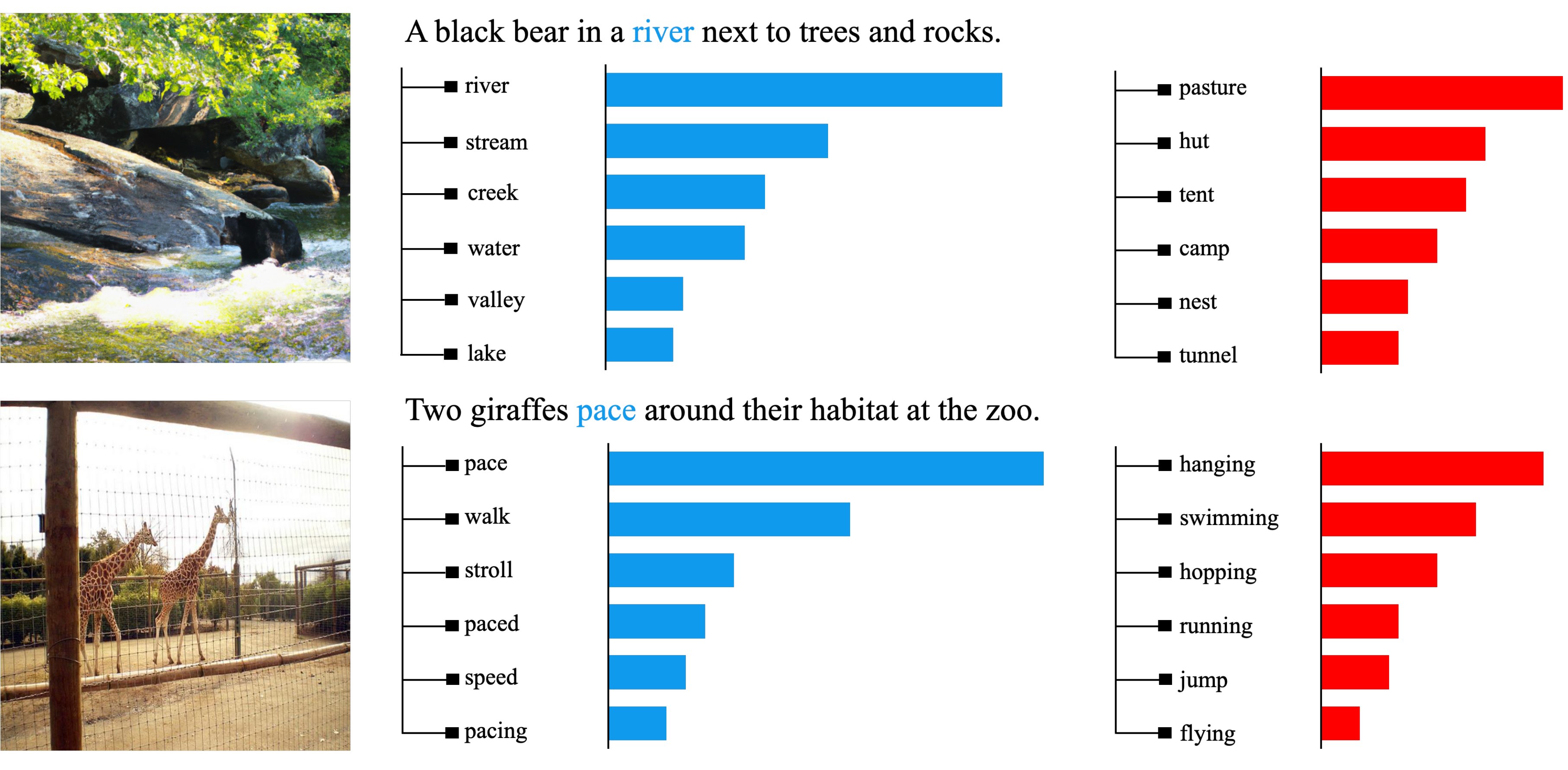}
    \caption{Examples of \smodel are presented in two columns: 1st col displays the soft labels generated by cross-distillation, while 2nd col lists the top negative words.} 
    \vspace{-0.4cm}
    \label{fig:case_probaility}
\end{figure}

\vpara{Discussion on Cross-Distillation.} 
The motivation behind \textit{cross-distillation} is rooted in the observation that when a complete caption without \texttt{[MASK]} token is fed into a language encoder, the encoder captures not only the contextual information surrounding each word but also information about the word itself. As shown in Figure~\ref{fig:case_probaility}, it allows the predicted tokens with high probability at each position to serve as potential synonyms or hypernyms of the original word, and the probability can be measured to determine the degree of similarity between the predicted words and the original word in context. This intuition ensures that substituting the masked token with the predicted token with a high probability will not alter the semantic meaning or grammatical structure of the sentence. Therefore, \textit{cross-distillation} is utilized to enhance the learning efficiency, representation, and generalization ability of the model. In this approach, we combine the one-hot labels of randomly masked ground-truth tokens with the soft labels generated by a frozen language encoder to train vision-language models. This enables smooth and efficient learning, where the potential synonyms generated by the frozen language encoder can be regarded as a variant of positive samples.

\subsection{Synthetic Hard Negatives for ITM} 

The purpose of Image-Text Matching is to capture the fine-grained alignment among image-text pairs. ITM can be viewed as a binary classification problem, which aims to predict whether an image-text pair is positive (matched) or negative (unmatched) based on the learned embeddings of the \texttt{[CLS]} token. Such embeddings are generated by the multimodal encoder, manifesting global cross-modality representations. Since the multimodal encoder adopts two cross-attention transformer modules, we concatenate two embeddings of the \texttt{[CLS]} token generated by vision and language modules respectively to obtain the final embeddings for ITM training. After that, a fully-connected (FC) layer with softmax activation function serves as a classifier to predict a two-class probability $p^{\text{itm}}$. The ITM loss can be represented as:
\begin{equation}
    \mathcal{L_{\text{itm}}} = \mathbb{E}_{(I,T) \sim D} H(y^{\text{itm}}, p^{\text{itm}}(I, T))
\end{equation}
where $y^{\text{itm}}$ denotes a binary ground-truth label.

A crucial method to improve performance on ITM is to find more informative negatives for model training. Negatives for ITM should be satisfied both of the following criteria:

\begin{itemize}
    \item \textit{The embedding of the hard negative sample should be similar to that of the positive sample in the embedding space. }
    \item \textit{The hard negative sample must be a true negative sample with some fine-grained features that contradict the positive sample.}
\end{itemize}

To accomplish the abovementioned criteria, we propose to synthesize hard negatives through textual augmentation, which differs significantly from ALBEF~\cite{li2021align} selecting negatives with higher contrastive similarities from ITC in the current batch. Such a synthetic method can generate negatives that are close to the positive sample in the embedding space while alleviating simultaneously  the false negative issue. Specifically, for any given positive image-text pair $(I, T)$, the synthetic hard negatives method generates its corresponding negative pair $(I, T_n)$, which can be divided into three steps: \\
 \indent 1) Generate the masked sentence $T'$: Substituting one word with a \texttt{[MASK]} tag in the original sentence $T$, based on part-of-speech (POS) tagging. The POS tagging helps identify crucial parts of a sentence, such as nouns, verbs, adjectives, and numerals. \\
\indent 2) Calculate predicted probability $p_m$: Taking the masked sentence $T'$ as input and passing it into the current language encoder to compute probabilities $p_m(t|T')$ at the \texttt{[MASK]} position during the beginning of each training step. \\
\indent 3) Synthesize hard negative sentence $T_n$: Sampling the probability of the corresponding \texttt{[MASK]} position $q_m^{\text{msk}}(T)$ from $q(T)$ which we have used in cross distillation module; calculating the negative sampling distribution $p_n = \frac{p_m(t|T')}{q_m^{\text{msk}}(T)}$; sampling one word based on $p_n$ to synthesize hard negative sentence $T_n$.

\vpara{Discussion on Hard Negatives.} In contrast to prior works that sample hard negatives from existing negatives, our proposed \smodel paradigm synthesizes negatives based on the current language encoder. As mentioned in the previous section, the plausibility of predicted words in the masked position can be measured by $p_m(t|T')$, while $q_m^{\text{msk}}(T)$ can gauge the similarity between the prediction words and the original word (See Figure~\ref{fig:case_probaility}). Consequently, selecting the predicted word with the highest $p_n$ ensures its plausibility in the context and its dissimilarity with the original word. This approach introduces a novel perspective on hard negatives, as it can be seen as a variant of data augmentation. Specifically, instead of generating positive samples for contrastive learning~\cite{he2020momentum,chen2020simple} through data augmentation, we aim to synthesize hard negative samples for the ITM objective in vision-language pretraining. These synthetic hard negatives offer additional information for model training since they conform to, but differ from, the positive samples, thereby accelerating the model's convergence and enhancing downstream performance.

\section{Experiments} \label{sec:experiment}
To demonstrate the effectiveness of \model, we conduct comprehensive experiments on 5 vision-language tasks. First, we introduce experimental setup, including model architecture, pre-training datasets, downstream tasks, and implementation details. Second, we compare our proposed \smodel with other classical vision-language pre-training models on various tasks, including visual question answering, visual reasoning, image-text retrieval, image captioning. Lastly, we design a series of ablation studies for model analysis.

\subsection{Experimental Setup}

\vpara{Pre-training Datasets.} In pre-training, we collect a vast number of image-text pairs from the Internet to train our model with two pre-training tasks (MLM and ITM). In line with previous research, we adopt four datasets for pre-training, including Conceptual Captions~\cite{sharma2018conceptual}, COCO~\cite{lin2014microsoft}, SBU Captions~\cite{ordonez2011im2text}, and Visual Genome~\cite{krishna2017visual}. Here, we utilize 4 million images for training since a large proportion of the image links have broken in the process of downloading datasets. Statistics of datasets are shown in Appendix A.

\vpara{Downstream Tasks.} We evaluate our proposed \smodel on 5 downstream vision-language tasks, including visual question answering, visual reasoning, visual entailment, image-text retrieval, and image captioning tasks. The detailed description of each task is represented in Appendix B. 

\begin{small}
    \begin{table*}[t!]
    \centering
    \renewcommand{\arraystretch}{1.15}
    \setlength{\tabcolsep}{2mm}
    \small
    {
    \begin{tabular}{ccccccccc}  
    \toprule
    \multirow{2}{*}{\bf Model} & \multirow{2}{*}{\textbf{\tabincell{c}{\#Pretrain \\ Images}}}  & \multirow{2}{*}{\textbf{\tabincell{c}{Visual \\ Encoder}}}   & \multicolumn{2}{c}{\bf VQAv2} & \multicolumn{2}{c}{\bf NLVR$^2$} & \multicolumn{2}{c}{\bf SNLI-VE} \\
    & & & \bf test-dev & \bf test-std &  \bf dev & \bf test & \bf dev  & \bf test  \\
    \midrule 
    \multicolumn{1}{c}{\it{BASE-Size Models}} \\
    \hline
    ViLT~\cite{kim2021vilt} & 4M & VIT-B-384/32 & 71.26 & - & 75.70 & 76.13 & - & - \\
    UNITER$_{\text{BASE}}$~\cite{chen2020uniter} & 4M & Faster R-CNN & 72.70 & 72.91 & 77.18 & 77.85 & 78.59 & 78.28 \\
    GLIPv2$_{\text{BASE}}$~\cite{zhang2022glipv2} & 20M & Swin-B-224 & 73.1 & 73.3 & - & - & - &\\
    VILLA$_{\text{BASE}}$~\cite{gan2020large}  & 4M & Faster R-CNN &73.59 & 73.67 & 78.39 & 79.30 & 79.47 & 79.03  \\
    UNIMO$_{\text{BASE}}$~\cite{li2020unimo}  & 4M & Faster R-CNN &73.79 & 74.02 & - & - & 80.00 & 79.10  \\
    CLIP-ViL$_{\text{BASE}}$~\cite{shen2021much} & 9.2M & CLIP-Res50 &73.92 & 74.09 & - & - &  78.64 & 78.97 \\
    KD-VLP~\cite{liu2021kd} & 4M & ResNet-101 & 74.20 & 74.31 & 77.36 & 77.78 & 78.21 & 77.87 \\
    ALBEF~\cite{li2021align} & 4M & DEIT-B-224/16 & 74.54 & 74.70 & 80.24 & 80.50 & 80.14 & 80.30 \\
    ALBEF~\cite{li2021align} & 14M& DEIT-B-224/16 &75.84 & 76.04 & 82.55 & 83.14  & 80.80 & 80.91 \\
    VinVL$_{\text{BASE}}$~\cite{zhang2021vinvl}  & 5.7M & ResNeXt-152 & 75.95 & 76.12 & 82.05 & 83.08 & - & -  \\
    VLMo$_{\text{BASE}}~\cite{bao2021vlmo}$ & 4M & MOME Transformer & 76.64 & 76.89 & 82.77 & 83.34  & - & -\\ 
    BLIP$_{\text{BASE}}$ ~\cite{li2022blip} & 14M & DEIT-B-224/16 & 77.54 & 77.62 & 82.67 & 82.30 & - & - \\
    METER-CLIP-ViT~\cite{dou2022empirical} & 4M & CLIP-ViT-B-224/16 & 77.68 & 77.64 & 82.33 & 83.05 & 80.86 & 81.19   \\ 
    SimVLM$_{\text{BASE}}$~\cite{wang2021simvlm} & 1.8B & ResNet-101 &77.87 & 78.14 & 81.72 & 81.77 & \bf 84.20  & \bf 84.15  \\
    OFA$_{\text{BASE}}$~\cite{wang2022ofa} & 54M & ResNet-101 &77.98 & 78.07 & - & -  & \demph{89.30$^\dagger$} & \demph{89.20$^\dagger$} \\
    X-VLM~\cite{zeng2021multi} & 4M & Swin-B-224 & 78.07 & 78.09 & \bf 84.16 & \underline{84.21} & - & - \\
    BLIP$_{\text{BASE}}$ ~\cite{li2022blip} & 129M & DEIT-B-224/16 & \underline{78.25} & \underline{78.32} & 82.15 & 82.24 & - & - \\
    \hline
    \hline
    \model$_{\text{BASE}}$ & 4M & CLIP-ViT-B-224/16 &\bf 78.62 & \bf 78.47 & \underline{83.21} & \bf 84.29 & \underline{81.50} & \underline{81.67} \\ 
    \hline
    \multicolumn{1}{c}{\it{Large-Size Models}} \\
    \hline
    UNITER$_{\text{LARGE}}$~\cite{chen2020uniter} & 4M & Faster R-CNN &73.82 & 74.02 & 79.12 & 79.98  & 79.39  & 79.38 \\
    VILLA$_{\text{LARGE}}$~\cite{gan2020large}  & 4M & Faster R-CNN &74.69 & 74.87 & 79.76 & 81.47 & 80.18 & 80.02  \\
    UNIMO$_{\text{LARGE}}$~\cite{li2020unimo}  & 4M & Faster R-CNN &75.06 & 75.27 & - & - & 81.11 & 80.63  \\
    VinVL$_{\text{LARGE}}$~\cite{zhang2021vinvl}  & 5.7M & ResNeXt-152 &76.52 & 76.60 & 82.67 & 83.98 & - & -  \\
    CLIP-ViL$_{\text{LARGE}}$~\cite{shen2021much} & 9.2M & CLIP-Res50$\times$4 &76.48 & 76.70 & - & - &  80.61 & 80.20 \\
    VLMo$_{\text{LARGE}}~\cite{bao2021vlmo}$ & 4M & MOME Transformer & \underline{79.94} & \underline{79.98} & \bf 85.64 & \bf 86.86  & - & -\\ 
    \hline
    \hline
    \model$_{\text{LARGE}}$ & 4M & CLIP-ViT-L-334/14 & \bf 80.19 & \bf 80.17 & \underline{85.16} & \underline{86.13} & \bf 83.12 & \bf 82.98 \\ 
    \hline
    \multicolumn{1}{c}{\demph{\it{Huge-Size Models}}} \\
    \hline
    \demph{SimVLM$_{\text{HUGE}}$~\cite{wang2021simvlm}} & \demph{1.8B} & \demph{Larger ResNet-152} & \demph{80.03} & \demph{80.34} & \demph{84.53} & \demph{85.15} & \demph{86.21} & \demph{86.32} \\
    \demph{BEIT-3~\cite{wang2022image}} & \demph{28M} & \demph{MOME Transformer} & \demph{84.19} & \demph{84.03} & \demph{91.51} & \demph{92.58} & \demph{-} & \demph{-} \\
    \demph{PaLI~\cite{chen2022pali}} & \demph{1.6B} & \demph{VIT-E-224} & \demph{84.30} & \demph{84.34} & \demph{-} & \demph{-} & \demph{-} & \demph{-} \\
    \bottomrule 
    \end{tabular}
    }
    \vspace{2mm}
    \caption{Result comparison with representative vision-language pre-training models. $^\dagger$ denotes using additional text premise as input.}
    \label{result}
    \end{table*}
\end{small}

\vpara{Implementation Details.} 
We train \smodel on 8 NVIDIA A100 GPUs for a total of 360,000 steps with a batch size of 1024, which takes a period of approximately 5 days. The maximum text sequence length is set to 50 and a max resolution of pre-training images is set to 288$\times$288. We utilize the AdamW optimizer~\cite{loshchilov2017decoupled} with an initial learning rate of $1e-5$ for the bottom vision and language encoders and $5e-5$ for the top multimodal encoder. To optimize the learning process, we adopt a linear decay learning rate schedule that contains a warm-up period with a ratio of 10$\%$. The learning rate is subsequently linearly decreased to $1e-8$ after 10$\%$ of the total training steps.

\subsection{Results on VL Classification Tasks}
We conduct an empirical evaluation of our proposed \smodel on vision-language (VL) classification tasks, including VQA, visual reasoning, and visual entailment. In order to demonstrate the effectiveness of \model, we compare it with SOTA methods and report the experimental results in Table~\ref{result}. It can be obviously observed that \smodel achieves impressive performances on the VQAv2 dataset and outperforms all baselines on both test-dev and test-std with either BASE or LARGE architecture. Notably, with the same amount of pre-training images (4M), \smodel significantly surpass other models~\cite{shen2021much, dou2022empirical} adopted CLIP-weights to initialize the vision encoder. Additionally, \smodel brings about  performance improvements in visual reasoning and visual entertainment over most of baselines, especially with the condition of the same amount of pre-training images. It clearly demonstrates that the superiority of our proposed \smodel in VL classification tasks. By performing BASE and LARGE architectures on the downstream classification tasks, the experimental results indicate that scaling the model's parameters can result in a significant performance improvements, which provides additional experimental supports to verify the effectiveness of \model.

\begin{table*}[htbp]
  \centering 
  \renewcommand{\arraystretch}{1.15}
    \setlength{\tabcolsep}{1.5mm}
    \small
    { 
    \begin{tabular}{l|cccccc|cccccc}
    \toprule
    \multirow{2}{*}{Method} & \multicolumn{6}{c|}{\textbf{Flickr}} & \multicolumn{6}{c}{\textbf{COCO}} \\
    & \bf IR@1 & \bf IR@5 & \bf IR@10 & \bf TR@1 & \bf TR@5 & \bf TR@10 & \bf IR@1 & \bf IR@5 & \bf IR@10 & \bf TR@1 & \bf TR@5 & \bf TR@10 \\
    \midrule
    ViLT$_{\text{BASE}}$~\cite{kim2021vilt} & 64.4 & 88.7 & 93.8 & 83.5 & 96.7 & 98.6 & 42.7  & 72.9 & 83.1 & 61.5 & 86.3 & 92.7 \\
    PixelBERT~\cite{huang2020pixel} &71.5 &92.1& 95.8 &87.0 &98.9 &99.5& 50.1 &77.6 &86.2 &63.6 &87.5 &93.6 \\
    UNITER$_{\text{BASE}}~\cite{chen2020uniter}$ & 72.5  & 92.4 & 96.1 & 85.9 & 97.1 & 98.8 & 50.3 & 78.5 & 87.2 & 64.4 & 87.4 & 93.1 \\
    VILLA$_{\text{BASE}}$~\cite{gan2020large} & 74.7 & 92.9 & 95.8 & 86.6 & 97.9 & 99.2 & - & - & - & - & - & - \\ 
    OSCAR~\cite{li2020oscar} & - & - & - & - & - & - & 54.0 & 80.8 & 88.5 & 70.0 & 91.1 & 95.5 \\
    VLMO$_{\text{BASE}}$~\cite{bao2021vlmo}  & 79.3 & 95.7 & 97.8 & 92.3 & 99.4 & 99.9 & 57.2 & 82.6 & 89.8 & 74.8 & 93.1 & \bf 96.9 \\
    ALBEF$_{\text{BASE}}$~\cite{li2021align} & 82.8 & 96.7 & 98.4& 94.3 & 99.4 & 99.8 & 56.8 & 81.5 & 89.2 & 73.1 & 91.4 & 96.0 \\
    METER-CLIP-ViT~\cite{dou2022empirical} & 82.2 & 96.3 & 98.3 & 94.3 & 99.6 & \bf 99.9 & 57.1 & 82.7 & \bf 90.0 & \bf 76.2 & \bf 93.2 & 96.8\\
    \hline
    \model$_{\text{BASE}}$ & \bf 85.2 & \bf 97.2 & \bf 98.8 & \bf 94.5 & \bf 99.8 & 99.8 & \bf 59.5 & \bf 83.1 & 89.7 &73.3 &91.8 &95.9 \\
    \bottomrule
    \end{tabular}
    }
    \caption{Experimental results on image retrieval (IR) and text retrieval (TR) on Flickr30K and COCO datasets.}
  \label{tab:retrieval_task}%
\end{table*}%

\begin{small}
\begin{table*}[htbp]
  \centering
   \renewcommand{\arraystretch}{1.15}
    \setlength{\tabcolsep}{1.6mm}
    \small
    {
    \begin{tabular}{l|cccc|cccc}
    \toprule
    \multirow{2}{*}{\bf Method} & \multicolumn{4}{c|}{\bf COCO} & \multicolumn{2}{c}{\bf NoCaps Val} & \multicolumn{2}{c}{\bf NoCaps Test} \\
    & \bf BLEU@4 & \bf METEOR &  \bf CIDEr & \bf SPICE & \bf CIDEr & \bf SPICE &  \bf CIDEr & \bf SPICE \\
    \midrule
    CLIP-ViL-ViT~\cite{shen2021much} & 21.1 & 19.4 & 58.0 & 12.2 & -& -& -&-\\
    GLIPv2$_{\text{BASE}}$~\cite{zhang2022glipv2} & 37.4 & - & 123.0 & 21.9 & - & - & - & - \\
    UFO$_{\text{BASE}}$~\cite{wang2021ufo} & 36.0 & 28.9 & 122.8 & 22.2 & 80.7 & 12.5 & 78.8 & 12.5 \\
    VinVL$_{\text{BASE}}$~\cite{zhang2021vinvl}  & 38.2 & 30.3 & 129.3 & 23.6 & 94.3$^*$ & 13.1$^*$ & 92.5$^*$ & 13.1$^*$\\
    METER-CLIP-ViT~\cite{dou2022empirical} & 38.8 & 30.0 & 128.2 & 23.0 & - & - & - & -  \\
    FIBER~\cite{dou2022coarse} & 39.1 & 30.4 & 128.4 & 23.1 & 88.6 & 13.0 & 86.0 & 12.9 \\
    SimVLM$_{\text{BASE}}$~\cite{wang2021simvlm} & 39.0 & 32.9 & 134.8 & 24.0 & - &- & - & -\\
    LEMON$_{\text{BASE}}$~\cite{hu2022scaling} &40.3 &30.2 &133.3 &23.3 & 100.4 & 13.8 & - & -\\
    \hline
    \model$_{\text{BASE}}$ & \bf 41.0 & \bf 30.9 & \bf 135.1 & \bf 23.6 & \bf 103.4 & \bf 13.9 & \bf 100.2 & \bf 13.9\\
    \bottomrule
    \end{tabular}}
    \vspace{2mm}
    \caption{Results on image captioning. $^*$ denotes that the model was optimized using the CIDEr metric to improve performance. It is worth noting that all the results reported for our model were obtained without CIDEr optimization.}
  \label{caption}

\end{table*}%
\end{small}

\subsection{Results on VL Retrieval Tasks}
Table~\ref{tab:retrieval_task} shows the results on VL retrieval tasks. We can find that \smodel outperforms existing VL models in most cases, especially on the Flickr dataset.
Notably, the results in recent studies~\cite{li2021align,li2022blip,dou2022coarse} illustrate the importance of ITC for retrieval tasks, which integrates ITC task into the pre-training and adopts a re-ranking strategy in the fine-tuning. Such investigations reveal that ITC can bring about a consistent improvement of 6\% in terms of R@1. The above fact may lead to slight performance improvements on the Flickr dataset and even performance degradation on the COCO dataset since \smodel only adopts MLM and ITM tasks for pre-training.

\subsection{Results on Image Captioning}
The experimental results in Table~\ref{caption} demonstrate that \smodel achieves better performance in the image captioning task. In specific, we fine-tune the model on the COCO Captions~\cite{lin2014microsoft} dataset and evaluate its performance with four metrics such as BLEU@4, METEOR, CIDEr, and SPICE. From Table~\ref{caption}, we can observe that \smodel consistently outperforms all baselines, especially compared to CLIP-ViL. Such results verify the effectiveness of \smodel rather than the importance of the initialization of CLIP-ViT. Furthermore, we conduct experiments on the NoCaps~\cite{agrawal2019nocaps} dataset without any additional fine-tuning or optimization techniques. It indicates that \smodel can urge the vision-language model to recognize fine-grained objects to enhance caption quality.

\subsection{Model Analysis}
To further analyze the impact of each component of \model, we perform a series of experiments with various objectives on VL classification and image captioning tasks. Table~\ref{ablation_study} demonstrate the performance comparison among various variants of our proposed \model. Compared to the basic variant (MLM), incorporating the ITM task can benefit all downstream vision-language tasks. Such benefits can be further enhanced with synthetic hard negatives and \textit{cross-distillation} method respectively. Synthetic hard negatives are utilized to boost the ITM tasks while \textit{cross-distillation} method is leveraged to improve the learning of the MLM task. We can find that the performance improvements of hard negatives are higher than \textit{cross-distillation}. By leveraging these two vital techniques, \smodel can achieve the best overall performance among different variants.


\begin{table}[htbp]
  \centering
  \vspace{2pt}
  \renewcommand{\arraystretch}{1.2}
  \setlength{\tabcolsep}{1.2mm}
  \small
  {
    \begin{tabular}{l|cccc}
    \toprule
   \bf Objectives & \bf VQAv2 & \bf NLVR2 & \bf SNLI-VE & \bf COCO$_{\text{Cap}}$\\
    \midrule
    MLM   &76.65  &82.21 & 80.16  & 38.2 \\
    MLM+ITM & 77.34 &82.97  & 80.90 & 38.8 \\
    MLM+ITM$_\text{{hard}}$ & 78.06 & 83.80  & 81.48 & 39.8 \\
    MLM$_{\text{distill}}$+ITM & 77.85  &83.52 & 81.22 & 39.6 \\
    \hline
   \model$_{\text{BASE}}$ & 78.47 & 84.29 & 81.67  & 40.4  \\
    \bottomrule
    \end{tabular}}
    \vspace{2mm}
    \caption{Impact of each component in \model.}
  \label{ablation_study}
\end{table}

\begin{figure*}[t!]
    \centering 
    \setlength{\abovecaptionskip}{5pt}
    \setlength{\belowcaptionskip}{-5pt}  \includegraphics[width=0.98\textwidth]{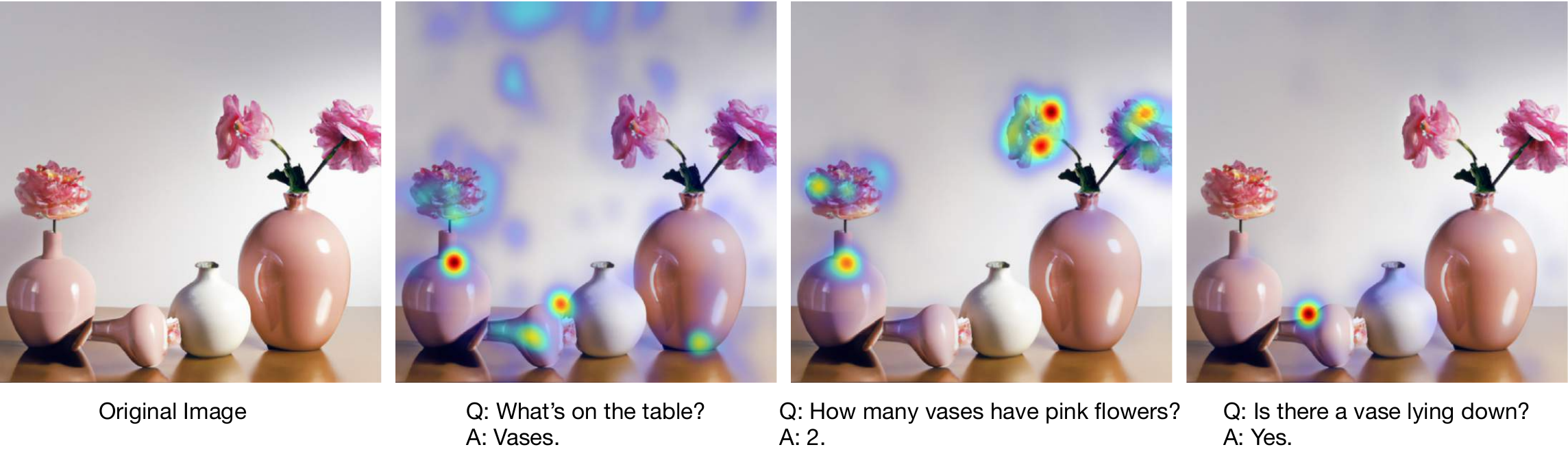}
    \caption{The Grad-CAM visualization of VQA.}
    \label{grad_vqa}
\end{figure*}

\subsection{Ablation Studies}
\vpara{Analysis on Image Captioning.}
We conduct an ablation study that aims to enhance the adaptability of pre-trained models for image captioning tasks in Table~\ref{caption_ablation}. Specifically, we introduce an additional pre-training phase that involved a language modeling (LM) task on the 4 million dataset for one epoch. Our results show that a improvement of 0.6 in BLEU@4 when incorporating the LM task. Furthermore, we investigate the issue of information leakage caused by deep fusion in the cross-attention modules. We explore two approaches to address this problem: removing the cross-attention module on top of the image encoder or transforming it into a self-attention module. We compare their performance and conclude that the cross-attention module should be retained for optimal performance.

\begin{small}
    \begin{table}[htbp]
        \centering
        \small
        \begin{tabular}{l|c|c|c|c}
        \toprule
            \multirow{2}{*}{\bf COCO} & \multicolumn{2}{c|}{\bf w/ LM} & \multicolumn{2}{c}{\bf w/o LM} \\
            &\bf w/o CA & \bf w/ CA & \bf w/o CA & \bf w/ CA \\
        \midrule
             BLEU@4 & 40.0 & 41.0 & 39.6 & 40.4\\
             CIDEr& 131.5 & 135.1 & 130.0 & 133.1\\ 
        \bottomrule
        \end{tabular}
        \vspace{2mm}
        \caption{Ablation study on image captioning. LM: language modeling pre-training. CA: cross-attention module. }
        \label{caption_ablation}
    \end{table}
\end{small}

\vpara{Impact of Negative Mining Method.} We conduct an experiment to analyze the impact of different negative mining methods. One straightforward approach is to utilize WordNet~\cite{miller1995wordnet} to generate negative samples by substituting antonyms from the original sentence. However, the experimental results in Table~\ref{negative_ablation} demonstrate that WordNet brings about a minimal performance improvement compared with in-batch randomly selected negative samples. Besides, pre-generated negatives are preprocessed at the beginning of pre-training, which performs better than WordNet but worse than dynamically-generated negatives. The possible reason is that dynamically-generated negatives are updated at each step based on the language encoder, ensuring the hardness of negative samples. More importantly, the combination of WordNet and dynamically-generated negatives shows performance degradation compared to the raw dynamically-generated negatives. Such finding can be attributed to the fact that the negatives generated by WordNet differ from the original sentence in the semantic space and representation space, making it easier for the model to distinguish them.

\begin{small}
    \begin{table}[htbp]
        \centering
        \begin{tabular}{l|c|c}
        \toprule
            \multirow{2}{*}{\bf Negative Mining Method} & \multicolumn{2}{c}{\bf VQAv2} \\
            & \bf test-dev & \bf test-std \\
        \midrule
            In-batch random & 74.08 & 74.37 \\
             WordNet & 74.24 & 74.36\\
             Pre-generated & 74.52 & 74.72 \\
             WordNet+Dynamically-generated & 74.71 & 74.86\\            
             \hline
             Dynamically-generated& \textbf{74.93} & \textbf{75.12}\\ 
        \bottomrule
        \end{tabular}
        \vspace{2mm}
        \caption{Ablation study on negative mining method. In the pre-generated approach, all negative samples are generated before the training phase, while in the dynamically-generated approach, negative samples for each batch are updated during each training step.}
        \vspace{-0.1cm}\label{negative_ablation}
    \end{table}
\end{small}

\subsection{Efficiency Analysis}

\smodel employs two distinct strategies to enhance downstream performance: the \textit{cross-distillation} technique in the Masked Language Modeling (MLM) task and the synthetic hard negative mining technique in the Information-Theoretic Metric (ITM) task. To shed light on the training cost associated with these techniques, we present the findings in Table~\ref{tab:time}. The combined implementation of both strategies introduces a marginal increase of approximately 10\% in training time per epoch. It is noteworthy, however, that this increment in training time is inconsequential in light of the benefits accrued. The adoption of these two techniques not only offsets the relatively minor increase in training duration but also contributes to the acceleration of model convergence.

\begin{table}[t!]
\begin{center}
\centering
\renewcommand{\arraystretch}{1.15}
    \setlength{\tabcolsep}{1.2mm}{
  \begin{tabular}{ccc}
    \toprule
  \multicolumn{2}{c}{\bf Pre-training Strategies} &  \multicolumn{1}{c}{\bf GPU-hours}\\
   \bf cross-distill \ \ \ & \bf hard negative &  \bf A100\\
    \midrule
     \XSolidBrush & \XSolidBrush  & 34.68 \\
    \Checkmark & \XSolidBrush  & 37.56  \\
    \XSolidBrush  & \Checkmark & 38.09  \\
    \Checkmark & \Checkmark & 38.43  \\
  \bottomrule
\end{tabular}}
\vspace{2mm}
\caption{The impact of different pre-training strategies on the time consumption of each epoch evaluated with a single A100 GPU.}
 \label{tab:time}
\end{center}
\vspace{-8mm}
\end{table}

\subsection{Case Study} 
\begin{figure*}
    \centering 
    \setlength{\abovecaptionskip}{5pt}
    \setlength{\belowcaptionskip}{-5pt}  \includegraphics[width=0.98\textwidth]{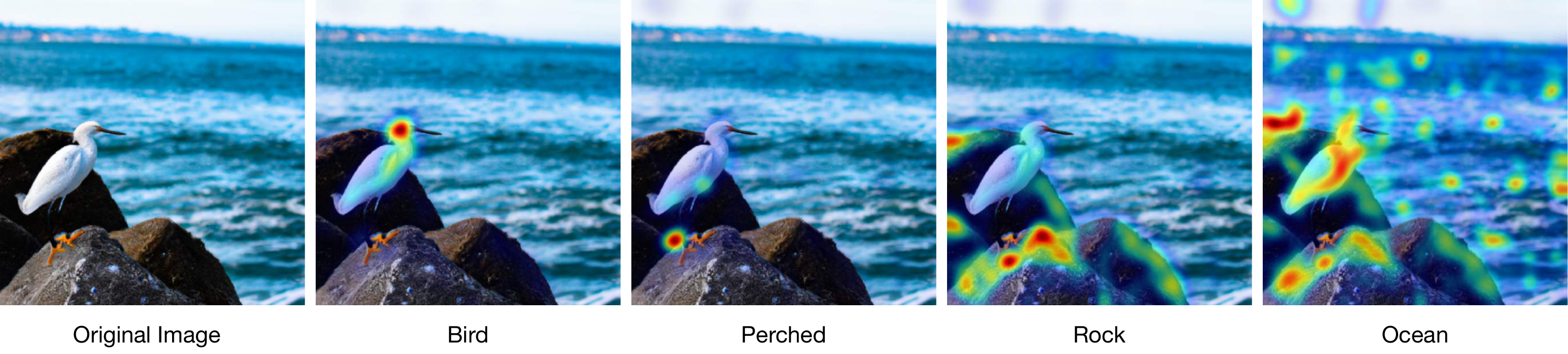}
    \caption{The Grad-CAM visualization of words in the caption “A white bird perched on a rock by the ocean.”}
    \label{grad_retrieval}
\end{figure*}

To demonstrate the effectiveness of \smodel in accurately identifying objects, attributes, actions, and quantitative relationships, we utilize Grad-CAM~\cite{selvaraju2017grad} for visualizations. Figure~\ref{grad_vqa} presents the results for the VQA task, indicating that our model can focus on fine-grained features in the image and correctly answer questions based on them. Besides, as shown in Figure~\ref{grad_retrieval}, visualizations for the retrieval task on a per-word basis illustrate the model's precise identification of objects and actions within the image, illustrating the model's precise identification of objects and actions within the image.

\section{Conclusion}  \label{sec:conclusion}

In this paper, we propose a novel vision-language pre-training method \smodel to further improve the representation ability of the model. Specifically, we propose a \textit{cross-distillation} method to generate soft labels to address the issue of treating synonyms of the masked words as negative samples in one-hot labels for MLM, which improves the robustness of vision-language models. Moreover, we design a new negative selection method for ITM, which aims to synthesize hard negatives based on the current language encoder by leveraging the context of language input. Such hard negatives provide more information for model convergence, which significantly enhances the downstream performances. Extensive experimental results on five vision-language tasks demonstrate the effectiveness and generalization ability of the proposed method.

\section{Acknowledgement}
This work is supported by the National Natural Science Foundation of China (No. 62277033). It also got partial support from National Engineering Laboratory for Cyberlearning and Intelligent Technology, and Beijing Key Lab of Networked Multimedia.

{\small
\bibliographystyle{ieee_fullname}
\bibliography{mybib}
}

\clearpage
\section*{Appendix}

\subsection*{A. Pre-training Details}

The statistics of pre-training datasets are presented in Table~\ref{dataset}. The COCO Captions dataset comprises manually generated captions where multiple captions are assigned to each image. For the Visual Genome dataset, the region description serves as the image caption, yielding several captions for each image. The SBU Captions and Conceptual Caption datasets contain a single caption per image. It should be noted that a considerable number of the image links in these two datasets have become invalid because they are collected from the Internet.

\begin{table}[htbp]
\centering
    \renewcommand{\arraystretch}{1.15}
    \setlength{\tabcolsep}{2mm}{
    \begin{tabular}{c|cccc}
    \toprule
     & \bf COCO  & \bf VG    & \bf SBU   & \bf CC3M  \\
    \midrule
    \#Images & 113K  & 108K  & 855K  & 2.98M  \\ 
    \hline
    \#Captions & 567K  & 5.4M  & 855K  & 2.98M  \\ 
    \bottomrule
    \end{tabular}}
    \vspace{2mm}
    \caption{Statistics of the pre-training datasets.}
  \label{dataset}
\end{table}

The default architecture of \smodel contains a dual-encoder architecture (a pre-trained vision encoder and a pre-trained language encoder) and a multimodal encoder. Table~\ref{tab:hyperparam_pretrain} reports the hyperparameters used in our pre-training model. For \model$_{\text{BASE}}$, we leverage a 12-layer transformer-based structure as language/vision encoder and 6-layer for the multimodal encoder respectively. The number of transformer layers for the language and vision encoders is set to 24 for \model$_{\text{LARGE}}$. The number of the multimodal encoder also maintains the default setup of 6-layer transformer-based structure. Here, we initialize the language encoder with weights from the pre-trained RoBERTa~\cite{liu2019roberta} and the vision encoder with weights from the pre-trained CLIP-ViT-224/16~\cite{radford2021learning}.

\begin{small}
\begin{table}[htbp]
\centering
    \renewcommand{\arraystretch}{1.15}
    \setlength{\tabcolsep}{0.35mm}{
    \begin{tabular}{c|cccc}
    \toprule
    \bf Hyperparameters & \bf ViLTA$_{\text{BASE}}$ & \bf ViLTA$_{\text{LARGE}}$ \\
      \midrule
      Total steps & $36$k & $24$k \\
      Warmup steps & $21.6$k & $14.4$k \\
      Batch size & $1024$ & $1024$ \\
      Learning rate & $1e^{-5}$ & $4e^{-6}$ \\
      Learning rate decay & \multicolumn{2}{c}{$\operatorname{Linear}$} \\
      Weight decay & \multicolumn{2}{c}{0.01} \\
      Dropout ratio & \multicolumn{2}{c}{$0.1$} \\
      AdamW $\epsilon$ & \multicolumn{2}{c}{$1e^{-8}$}  \\
      AdamW $\beta$ & \multicolumn{2}{c}{(0.9, 0.98)} \\
      \hline
      Textual encoder & RoBERTa$_{\text{BASE}}$ & RoBERTa$_{\text{LARGE}}$ \\
      Visual encoder & CLIP-ViT-B-224 & CLIP-ViT-L-336 \\
      Patch size & $16$ & $14$ \\
      Input resolution & $288$ & $224$ \\
      \hline
    Number of layers & $6$ & $6$ \\
      Hidden size & $768$ & $1024$ \\
      FFN inner hidden size & $3072$ & $4096$ \\
      Number of attention heads & $12$ & $16$ \\
    \bottomrule
    \end{tabular}}
    \vspace{2mm}
    \caption{
     Hyperparameters for pre-training model. The last block is the hyperparameters for the multimodal encoder.
  }
  \label{tab:hyperparam_pretrain}
\end{table}
\end{small}

\subsection*{B. Fine-tuning Details} 
We fine-tune \smodel on 5 downstream tasks using the hyperparameters reported in Table~\ref{tbl:ft:vqa_nlvr2:hyperparams} for VL classification tasks, Table~\ref{tbl:ft:retrieval:hyperparams} for VL retrieval tasks, Table~\ref{tbl:ft:captioning:hyperparams} for image captioning. In the following sections, we provide a comprehensive description of the fine-tuning configurations employed for each task.
\begin{itemize}
    \item \textit{Visual Question Answering (VQA)}~\cite{antol2015vqa} aims to predict a natural language answer based on the given image and question. Following the previous works~\cite{kim2021vilt,dou2022empirical,bao2021vlmo,wang2022image}, we treat VQA as a multi-label classification task with 3,129 possible answers. We concatenate the image representation $v_{cls}$ and text representation $w_{cls}$ obtained from the multimodal model, and then pass it through a 2-layer MLP layer to perform a classification task. We use GELU activation function and a binary cross-entropy loss function on the soft target scores to optimize the model.

    \item \textit{Visual Reasoning} focuses on predicting whether the caption is true or false for a pair of images. Here, we employ a pairwise strategy to effectively process the input in NLVR$^2$~\cite{suhr2018corpus} dataset, where each data sample is divided into \textit{(image1, statement)} and \textit{(image2, statement)}. We then feed them separately into the model to obtain two representations and concatenate them together to pass through a binary classification head.

     \item \textit{Visual Entailment} aims to predict whether a natural language hypothesis is entailed, neutral or contradicted by the image premise. We train and evaluate our model on SNLI-VE~\cite{xie2019visual} dataset and treat it as a three-class classification problem.
    
    \item \textit{Image-Text Retrieval} contains two sub tasks: image-to-text retrieval (TR) and text-to-image retrieval (IR). COCO~\cite{lin2014microsoft} and Flickr30K~\cite{plummer2015flickr30k} serve as evaluation datasets. Following the standard setting in ViLT~\cite{kim2021vilt}, We use the pre-trained ITM head, specifically the component that calculates the true-pair logits, to initialize the similarity score head. We then sample 15 random texts as negative examples and use a cross-entropy loss that maximizes the scores for positive pairs.

    \item \textit{Image Captioning} is a generative task and we investigate whether our encoder-only model is suitable for such generative tasks. To adapt our model for image captioning, we modify the encoder on the text side of the model by transforming it into a causal decoder through the adjustment of the attention mask. Subsequently, we fine-tune the model on the COCO Captions~\cite{lin2014microsoft} dataset using cross-entropy loss and evaluate it on the NoCaps~\cite{agrawal2019nocaps} dataset without additional training. 
\end{itemize}

\begin{table}[h]
\centering
\renewcommand{\arraystretch}{1.15}
\setlength{\tabcolsep}{1.2mm}{
\begin{tabular}{l|ccc}
\toprule
\bf Hyperparameters & \bf VQAv2 & \bf NLVR$^2$ & \bf SNLI-VE \\ 
\midrule
Learning rate & $1e^{-5}$ & $1e^{-5}$ & $2e^{-6}$\\
Epochs & 10  & 10 & 5\\
Batch size & 512 & 256  & 64\\
AdamW $\epsilon$ & \multicolumn{3}{c}{$1e^{-8}$}  \\
AdamW $\beta$ & \multicolumn{3}{c}{(0.9, 0.98)} \\
Weight decay & 0.05 & 0.01 & 0.01 \\
Dropout ratio & \multicolumn{3}{c}{0.1} \\
Input resolution & $576^2$ & $384^2$ & $288^2$\\
\bottomrule
\end{tabular}}
\vspace{2mm}
\caption{
Hyperparameters for fine-tuning ViLTA on VL classification tasks.
}
\label{tbl:ft:vqa_nlvr2:hyperparams}
\end{table}

\begin{table}[h]
\centering
\renewcommand{\arraystretch}{1.15}
\setlength{\tabcolsep}{1.2mm}{
\begin{tabular}{l|cc}
\toprule
\bf Hyperparameters & \bf COCO  & \bf Flickr\\
\midrule
Learning rate & \multicolumn{2}{c}{$5e^{-6}$} \\
Epochs & \multicolumn{2}{c}{$10$} \\
Batch size & \multicolumn{2}{c}{$64$} \\
AdamW $\epsilon$ & \multicolumn{2}{c}{$1e^{-8}$}  \\
AdamW $\beta$ & \multicolumn{2}{c}{$(0.9, 0.98)$} \\
Weight decay & \multicolumn{2}{c}{$0.01$} \\
Dropout ratio& \multicolumn{2}{c}{$0.1$} \\
Input resolution & \multicolumn{2}{c}{$576^2$} \\
\bottomrule
\end{tabular}}
\vspace{2mm}
\caption{
Hyperparameters for fine-tuning ViLTA on VL retrieval tasks.
}
\label{tbl:ft:retrieval:hyperparams}
\end{table}

\begin{table}[h]
\centering
\renewcommand{\arraystretch}{1.15}
\setlength{\tabcolsep}{1.2mm}{
\begin{tabular}{l|c}
\toprule
\bf Hyperparameters & \bf COCO Captioning \\
\midrule
Learning rate & $1e^{-5}$ \\
Epochs & 10 \\
Batch size & 512 \\
AdamW $\epsilon$ & $1e^{-8}$  \\
AdamW $\beta$ & (0.9, 0.98) \\
Weight decay & 0.01 \\
Dropout ratio& 0.1 \\
Input resolution & $576^2$ \\
Label smoothing $\varepsilon$ & 0.1 \\
Beam size & 5 \\
\bottomrule
\end{tabular}}
\vspace{2mm}
\caption{
Hyperparameters for fine-tuning \smodel on image captioning.
}
\label{tbl:ft:captioning:hyperparams}
\end{table}

\subsection*{C. Scaling Ability} 
To show the effectiveness of ViLTA on extensive datasets, we expand the training of ViLTA-base and ViLTA-large on a subset of the LAION-2B and CC12M datasets employing 64 A100 GPUs in Table~\ref{tab:retrieval_task}. The total volume of data was roughly 150M, comparable to the 129M dataset used in BLIP. All performance metrics for retrieval tasks show substantial enhancements, ranging from 73.3 to 80.5 on the COCO dataset for text retrieval in terms of recall@1. However, the gain in VL understanding (VLU) tasks is not as prominent as the increase in retrieval tasks, which is consistent with the findings in previous studies~\cite{li2022blip, bugliarello2023measuring}. Such discrepancy arises due to the challenges associated with the considerable noise present in large-scale web data, which are integral to VLU tasks. As shown in Table~\ref{nlvr}, in the context of a large-scale dataset, ViLTA achieves a better gain, while, in contrast, BLIP brings about performance degradation. 

\begin{table}[htbp]
  \centering 
  \renewcommand{\arraystretch}{1.15}
    \setlength{\tabcolsep}{1.2mm}
    \small
    { 
    \begin{tabular}{l|ccc|ccc}
    \toprule
    \multirow{2}{*}{\bf Dataset} & \multicolumn{3}{c|}{\textbf{Flickr}} & \multicolumn{3}{c}{\textbf{COCO}} \\
    & \bf TR@1 & \bf TR@5 & \bf TR@10  & \bf TR@1 & \bf TR@5 & \bf TR@10 \\
    \midrule
    \bf 4M & 94.5 & 99.8  & 99.8& 73.3 & 91.8 & 95.9  \\
    \hline  
    \bf 150M & \bf 95.7 & \bf 99.9 & \bf 99.9 & \bf 80.5 & \bf 94.6 & \bf 97.3  \\
    \bottomrule
    \end{tabular}
    }
    \caption{Experimental results on  retrieval task.}
  \label{tab:retrieval_task}%
\end{table}%

\begin{small}
    \begin{table}[h]
        \centering
        \begin{tabular}{l|cc|cc}
        \toprule
            \multirow{2}{*}{\bf Dataset} & \multicolumn{2}{c|}{\bf BLIP} & \multicolumn{2}{c}{\bf ViLTA} \\
            &\bf 14M  & \bf 129M  & \bf 4M  & \bf 129M  \\
        \midrule
             \bf NLVR2-dev & 82.67 & 82.15 & 85.16 & 86.33\\
             \midrule
             \bf NLVR2-test & 82.30 & 82.24 & 86.13 & 87.25\\ 
        \bottomrule
        \end{tabular}
        \vspace{2mm}
        \caption{Results on NLVR2 dataset. Large scale data may not have significant benefits for VLU tasks. }
        \label{nlvr}
    \end{table}
\end{small}

\subsection*{D. Additional Results} 
\begin{figure*}[t!]
    \centering 
    \setlength{\abovecaptionskip}{5pt}
    \setlength{\belowcaptionskip}{-5pt}  \includegraphics[width=0.98\textwidth]{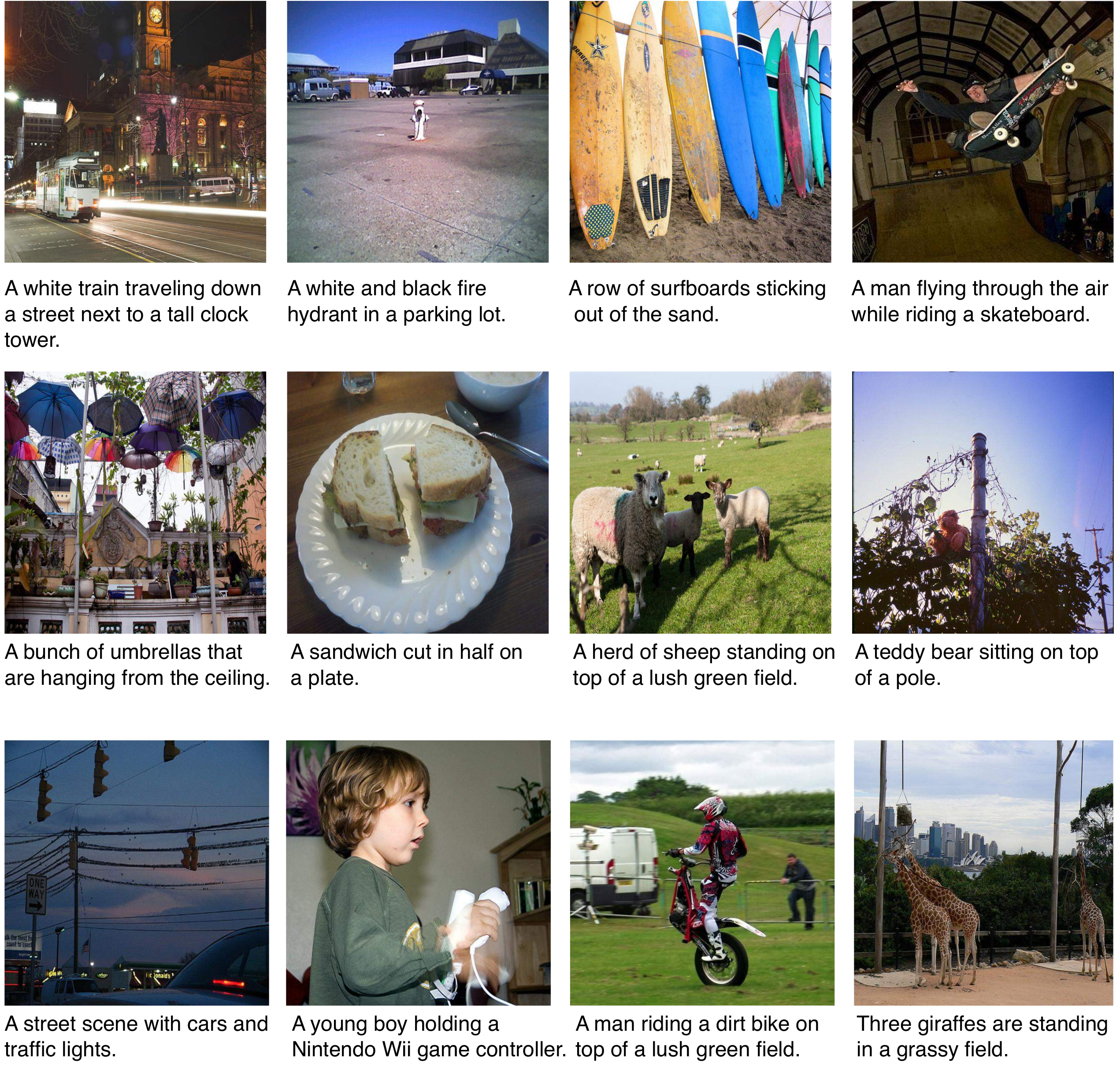}
    \caption{Case study of \smodel on image captioning task.}
    \label{captions}
\end{figure*}

In this section, we present additional results generated by \model. Specifically, we show the efficacy of \smodel in image captioning. The case study in Figure~\ref{captions} shows the generated image captions on a series of samples. Notably, \smodel generates diverse and descriptive captions, which can effectively encapsulate the content of the corresponding images. These results verify the effectiveness of \smodel in different VL tasks.

\end{document}